\documentclass[conference]{IEEEtran}
\IEEEoverridecommandlockouts

\usepackage{cite}
\usepackage{amsmath,amssymb,amsfonts}
\usepackage{algorithmic}
\usepackage{graphicx}
\usepackage{textcomp}
\usepackage{xcolor}
\usepackage{hyperref}

\def\BibTeX{{\rm B\kern-.05em{\sc i\kern-.025em b}\kern-.08em
    T\kern-.1667em\lower.7ex\hbox{E}\kern-.125emX}}
\begin{document}
\title{HyperFake: Hyperspectral Reconstruction and Attention-Guided Analysis for Advanced Deepfake Detection}

\author{\IEEEauthorblockN{Pavan C Shekar\IEEEauthorrefmark{1}, Pawan Soni\IEEEauthorrefmark{1}, Vivek Kanhangad\IEEEauthorrefmark{1}}
\IEEEauthorblockA{\IEEEauthorrefmark{1}Department of Electrical Engineering, Indian Institute of Technology Indore, Indore, 453552, India\\
Email: \{ee200002059, phd2201102014, kvivek\}@iiti.ac.in}
}

\maketitle

\begin{abstract}
Deepfakes pose a significant threat to digital media security, with current detection methods struggling to generalize across different manipulation techniques and datasets. While recent approaches combine CNN-based architectures with Vision Transformers or leverage multi-modal learning, they remain limited by the inherent constraints of RGB data. We introduce HyperFake, a novel deepfake detection pipeline that reconstructs 31-channel hyperspectral data from standard RGB videos, revealing hidden manipulation traces invisible to conventional methods. Using an improved MST++ architecture, HyperFake enhances hyperspectral reconstruction, while a spectral attention mechanism selects the most critical spectral features for deepfake detection. The refined spectral data is then processed by an EfficientNet-based classifier optimized for spectral analysis, enabling more accurate and generalizable detection across different deepfake styles and datasets, all without the need for expensive hyperspectral cameras. To the best of our knowledge, this is the first approach to leverage hyperspectral imaging reconstruction for deepfake detection, opening new possibilities for detecting increasingly sophisticated manipulations. Code available at: \url{https://github.com/pavan98765/HyperFake}.

\end{abstract}


\section{Motivation}

The rapid advancement of deepfake technology has introduced a profound challenge to digital security and trust. Every day, thousands of manipulated videos flood the internet, ranging from harmless entertainment to dangerous misinformation capable of influencing politics, media, and cybersecurity. While early deepfakes exhibited visible artifacts that made detection relatively straightforward, modern AI-driven methods—powered by generative adversarial networks (GANs) and diffusion models—produce hyper-realistic fakes that can deceive both human perception and state-of-the-art detection algorithms.

Traditional deepfake detection methods rely primarily on RGB (red, green, blue) color information, similar to what a standard camera captures. However, this limited spectral representation fails to reveal the subtle, imperceptible inconsistencies introduced during manipulation. Just as an art authenticator requires specialized tools beyond the naked eye to detect forged paintings, deepfake detectors need richer spectral information beyond standard RGB inputs to uncover hidden manipulation traces.

To address this challenge, we introduce HyperFake, a novel deepfake detection pipeline that leverages hyperspectral reconstruction to extract fine-grained spectral details from RGB videos. By reconstructing 31-channel hyperspectral data using an improved MST++ model, HyperFake unveils manipulation artifacts invisible in conventional RGB space. This approach enables more robust and generalizable deepfake detection, without the need for specialized hyperspectral cameras, making it both accessible and scalable for real-world applications.

\section{The Power of Hyperspectral Imaging}
Traditional RGB cameras capture images using only three color channels—red, green, and blue—limiting their ability to detect subtle artifacts in manipulated media. In contrast, hyperspectral imaging (HSI) captures information across dozens of spectral bands, revealing fine-grained material properties that are invisible to the human eye. This technology has been widely used in fields such as remote sensing, medical imaging, and forensic analysis, where deeper spectral insights provide crucial advantages over standard imaging techniques.

HSI has already demonstrated its effectiveness in medical imaging, where it can differentiate between healthy and diseased tissue, even when both appear identical in standard RGB images. Similarly, deepfake artifacts—such as unnatural lighting effects, altered textures, and spectral inconsistencies—may remain hidden in RGB but become detectable with hyperspectral analysis.

A useful analogy is currency verification: just as security features invisible under normal light become apparent under UV or infrared illumination, hyperspectral imaging can reveal deepfake manipulation traces that RGB-based methods fail to detect. By analyzing subtle spectral variations, HSI provides an additional layer of security in identifying AI-generated forgeries.

Despite its potential, the biggest challenge in applying HSI to deepfake detection has been the high cost of dedicated hyperspectral sensors, which can range from tens to hundreds of thousands of dollars. This makes real-world adoption impractical. To overcome this barrier, our approach reconstructs hyperspectral data from standard RGB inputs, enabling deepfake detection without the need for expensive specialized hardware.\cite{Nirkin2020}

\subsection{Our Solution: HyperFake}
To overcome this barrier, we introduce HyperFake, a deepfake detection pipeline that reconstructs high-fidelity hyperspectral data from standard RGB videos. Instead of relying on expensive hyperspectral cameras, HyperFake leverages an enhanced MST++ model to estimate 31-channel spectral information from conventional RGB inputs.

By reconstructing this rich spectral representation, HyperFake reveals manipulation artifacts invisible in standard RGB space, significantly improving deepfake detection robustness. Furthermore, we integrate a spectral attention mechanism to refine spectral features, ensuring that only the most discriminative information is utilized for classification. This approach makes hyperspectral analysis accessible and scalable, enabling high-accuracy deepfake detection without requiring specialized hardware.

\subsection{Contributions}
\begin{itemize}
    \item \textbf{Hyperspectral-enhanced deepfake detection}: First pipeline to reconstruct 31-channel hyperspectral data from standard RGB, revealing manipulation artifacts invisible in conventional color space.
    \item \textbf{Spectral attention for improved feature extraction:} Novel spectral attention mechanism that identifies and enhances manipulation-sensitive spectral bands, improving classification accuracy.
    \item \textbf{Robust and scalable detection:} Optimized EfficientNet-based classifier with spectral recalibration layers, validated on deepfake dataset for improved generalization without specialized hardware.
\end{itemize}

\section{Related Work}
The rapid evolution of deepfake technology has led to the development of various detection strategies, ranging from CNN-based feature extraction to transformer-based models. While these methods improve detection accuracy, they often struggle with generalization across datasets and deepfake types. Multi-modal and self-supervised approaches aim to address this issue, but they remain reliant on RGB-based information, which lacks the spectral depth needed to expose subtle manipulation artifacts. This section reviews traditional deepfake detection methods and highlights the emerging role of spectral analysis in media forensics, setting the foundation for our proposed hyperspectral reconstruction approach.

\subsection{Evolution of Deepfake Detection}
Traditional deepfake detection relied on CNN-based models that analyzed pixel-level anomalies, such as unnatural facial movements, mismatched shadows, and color inconsistencies. While effective against early deepfakes, these models struggled with generalization. As generative models improved, deepfake artifacts became increasingly imperceptible to standard CNNs, reducing detection accuracy on unseen manipulations. To mitigate this, researchers integrated CNNs with Vision Transformers (ViTs), leveraging the strength of CNNs for localized feature extraction and ViTs for capturing long-range dependencies. This hybrid approach showed promise but still suffered from reliance on RGB features, which are often insufficient for detecting advanced deepfakes.

To enhance robustness, recent strategies have integrated CNNs with Vision Transformers (ViTs)\cite{Coccomini2022}, capitalizing on the strengths of both architectures. ViTs, known for their capability to model long-range dependencies, complement the localized feature extraction of CNNs. This hybrid approach has shown promise in capturing both local and global features, thereby improving detection accuracy. Despite these advancements, achieving reliable detection across diverse scenarios remains a significant challenge, as deepfake generation techniques continue to evolve rapidly.\cite{laanet}

\subsection{Multi-Modal and Self-Supervised Approaches}
To improve generalization and reduce reliance on large labeled datasets, researchers have explored multi-modal learning, which combines information from various data sources such as audio, video, and textual data. For instance, the FakeOut pipeline\cite{fakeout} demonstrated that leveraging multiple data modalities enhances the model's ability to detect deepfakes across various contexts. By analyzing the coherence between audio and lip movements or the consistency between textual content and visual cues, multi-modal approaches can identify discrepancies indicative of deepfakes.

In addition to multi-modal learning, self-supervised learning techniques have been employed to learn more generalized features. These methods involve pre-training models on large amounts of unlabeled data to capture underlying data structures, which can then be fine-tuned for specific tasks like deepfake detection. Self-supervised approaches have shown promise in capturing subtle patterns and anomalies, thereby enhancing detection capabilities without the need for extensive labeled datasets.

\subsection{Spectral Analysis in Media Forensics}
Beyond traditional RGB imaging, spectral analysis has been applied in media forensics to detect forgeries. Hyperspectral Imaging (HSI) captures information across a wide range of wavelengths, providing detailed spectral signatures for each pixel in an image. This technology has been effectively utilized in fields such as art conservation and forensic document examination, where it aids in identifying subtle material differences and uncovering alterations invisible to the naked eye.

In forensic document examination, HSI has been used to differentiate between inks that appear identical under standard lighting conditions but exhibit distinct spectral properties. Similarly, in art conservation, HSI assists in revealing underlying sketches, differentiating between original and restored areas, and identifying pigments used by artists. These applications highlight the potential of spectral analysis in uncovering hidden details and verifying authenticity.

However, the application of HSI in deepfake detection has been limited, primarily due to the high cost and complexity of hyperspectral cameras. Recent advancements aim to reconstruct hyperspectral data from standard RGB inputs, potentially overcoming these barriers and providing new avenues for detecting sophisticated forgeries. By estimating hyperspectral information from conventional RGB images, it becomes possible to analyze spectral discrepancies introduced during the deepfake generation process, thereby enhancing detection capabilities\cite{blockwise, deepfakeReview}.

\begin{figure*}[!t]
\centering
\includegraphics[width=\textwidth]{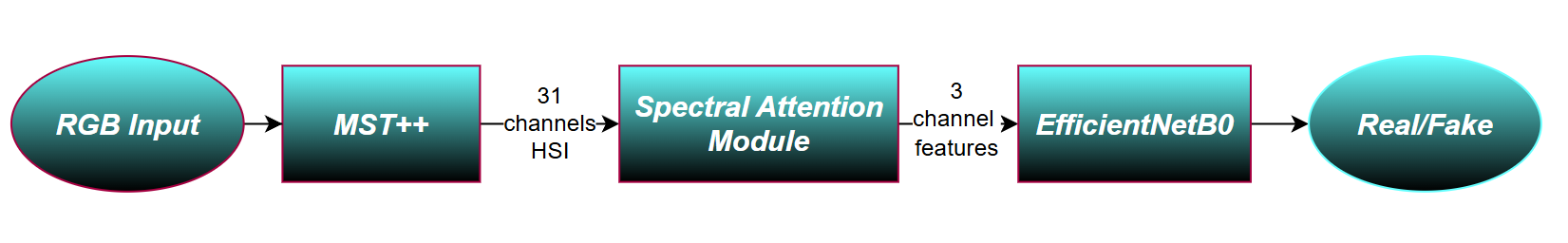}
\caption{HyperFake pipeline architecture for Deepfake detection. RGB input is processed by MST++ to generate a 31-channel hyperspectral image. The spectral attention
module selects key features, reducing them to three channels. EfficientNetB0 then classifies the input as real or fake.}
\label{fig:pipeline}
\end{figure*}

\begin{figure}[!t]
\centering
\includegraphics[width=\linewidth]{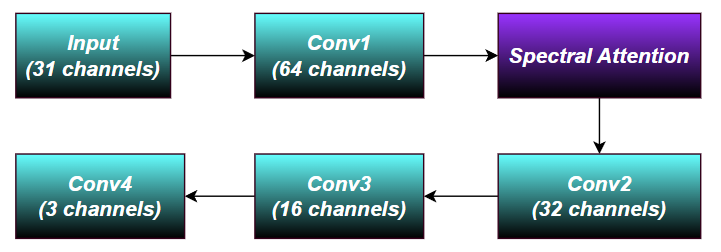}
\caption{Spectral Attention Module: Transforming 31-Channel Hyperspectral Data to 3-Channel Representation
Diagram illustrating the spectral attention mechanism that identifies and emphasizes the most critical spectral features, reducing dimensionality while preserving key discriminative information for deepfake detection.}
\label{fig:spectral_to_rgb}
\end{figure}

\section{Methodology}

\subsection{Overview}
The \textbf{HyperFake} pipeline enhances deepfake detection by leveraging \textbf{hyperspectral reconstruction}, \textbf{spectral attention mechanisms}, and an \textbf{efficient classification network}. It consists of three key modules:

\begin{itemize}
    \item \textbf{Hyperspectral Reconstruction Module}: Converts RGB inputs into \textbf{31-channel hyperspectral data} using an enhanced MST++ model.
    \item \textbf{Spectral Attention Module}: Identifies and enhances the most discriminative spectral features for deepfake detection.
    \item \textbf{Classification Module}: Uses an EfficientNetB0-based network\cite{efficientnet}, optimized with spectral-aware recalibration, to classify real vs. fake videos.
\end{itemize}

By seamlessly integrating these components, \textbf{HyperFake} enhances \textbf{generalization across diverse deepfake types} and significantly improves \textbf{detection accuracy}.

\subsection{Hyperspectral Image Reconstruction}
The \textbf{first stage} of HyperFake reconstructs \textbf{hyperspectral information} from standard RGB inputs using an improved \textbf{MST++ model}\cite{MSTPlusPlus2022}. Hyperspectral data, which contains \textbf{31 spectral bands}, provides rich material-specific information that helps uncover \textbf{deepfake artifacts invisible in RGB space}.

\subsubsection{MST++ for Hyperspectral Estimation}
MST++ is a multi-stage spectral-wise transformer model designed to reconstruct hyperspectral information from RGB images. It leverages a combination of advanced attention mechanisms to enhance spectral feature extraction and improve reconstruction accuracy.

The model employs Spectral-wise Multi-head Self-Attention (S-MSA) to capture correlations between spectral bands, ensuring that dependencies across different wavelengths are effectively modeled. Additionally, Spectral-wise Attention Blocks (SABs) refine spectral feature extraction by selectively enhancing the most relevant spectral information while reducing noise. To further maintain spectral coherence, Single-stage Spectral-wise Transformers (SSTs) are incorporated to improve spectral consistency, ensuring that reconstructed hyperspectral representations accurately reflect the underlying material properties.

To optimize MST++ for deepfake detection, we introduce FlexiAttention, a novel mechanism that enhances hyperspectral reconstruction by downsampling spectral features, applying attention, and then upsampling them. This approach allows the model to focus on the most discriminative spectral regions while reducing computational overhead. By selectively refining spectral information during reconstruction, FlexiAttention improves both reconstruction quality and processing speed, leading to a PSNR increase from 34.32 dB in the baseline MST++ model to 34.8 dB. This enhancement ensures higher spectral fidelity, making deepfake artifacts more distinguishable. Additionally, by reducing redundant computations through efficient attention-based feature selection, FlexiAttention accelerates inference, making MST++ more practical for large-scale deepfake detection.

\begin{equation}
H = R(I_{RGB}) \in \mathbb{R}^{B\times31\times H\times W}
\end{equation}

where \( H \) is the reconstructed hyperspectral representation of the input RGB image \( I_{RGB} \), covering \textbf{31 spectral bands}.

\subsubsection{Training for Hyperspectral Reconstruction}
To ensure high-fidelity spectral reconstruction, we \textbf{train the improved MST++ from scratch} on the \textbf{ARAD 1K dataset}\cite{arad1k}, which contains extensive hyperspectral data. The pretrained MST++ weights are then \textbf{frozen and integrated} into the HyperFake pipeline, allowing the model to focus on \textbf{deepfake classification} while leveraging hyperspectral features.

\subsection{Spectral Attention Mechanism}
After reconstructing \textbf{31-channel hyperspectral data}, we apply a \textbf{Spectral Attention Mechanism} to \textbf{refine and extract the most discriminative spectral features}. This module \textbf{selects key spectral bands that exhibit strong deepfake-related artifacts}, such as unnatural light reflections and color inconsistencies.

\subsubsection{Attention-Based Feature Selection}
\begin{equation}
SA(X) = \text{softmax}\left( \frac{(X W_Q)(X W_K)^T}{\sqrt{d}} \right)(X W_V)
\end{equation}

where:
\begin{itemize}
    \item \( X \) is the input hyperspectral feature map.
    \item \( W_Q, W_K, W_V \) are query, key, and value matrices.
    \item The \textbf{softmax operation} ensures that highly relevant spectral bands receive greater attention.
\end{itemize}

\subsubsection{Dimensionality Reduction for Efficient Processing}
While hyperspectral data consists of \textbf{31 channels}, many CNN-based classifiers operate on \textbf{3-channel inputs}. The Spectral Attention Module reduces hyperspectral data into an optimized \textbf{3-channel representation} by \textbf{selecting the most informative spectral bands}.

\begin{equation}
F_{reduced} = \sum_{i=1}^{31} \alpha_i H_i
\end{equation}

where \( \alpha_i \) are learned attention weights that emphasize \textbf{manipulation-sensitive spectral bands}, reducing noise and improving generalization.

\subsection{Loss Function}

To ensure \textbf{accurate classification} in deepfake detection, we use \textbf{Binary Cross-Entropy with Logits Loss (BCEWithLogitsLoss)} as our primary loss function. The classification loss is defined as:

\begin{equation}
L_{cls} = -\frac{1}{N} \sum_{i=1}^{N} \left[ y_i \log(\sigma(z_i)) + (1 - y_i) \log(1 - \sigma(z_i)) \right]
\end{equation}

where:
\begin{itemize}
    \item \( N \) is the number of samples in a batch.
    \item \( z_i \) represents the raw model output (logit) for the \( i \)-th sample.
    \item \( y_i \) is the ground truth label (either 0 or 1).
    \item \( \sigma(z_i) = \frac{1}{1 + e^{-z_i}} \) is the \textbf{sigmoid activation function}, which maps logits to probabilities.
\end{itemize}

Unlike standard Binary Cross-Entropy (BCE) loss, \textbf{BCEWithLogitsLoss applies the sigmoid activation internally}, improving numerical stability and preventing issues like vanishing gradients. This approach ensures that the model efficiently distinguishes between real and fake deepfakes while maintaining robustness during evaluation.

\subsection{Classification Network}
The final classification stage uses EfficientNet-B0, a lightweight yet powerful convolutional neural network for deepfake detection. EfficientNet-B0 is designed to prioritize important features while suppressing irrelevant ones, improving detection accuracy. Its ability to capture both local and global patterns helps enhance generalization across unseen deepfake datasets, making it a reliable choice for classification.

\subsection{Training Protocol}
We train HyperFake using the Adam optimizer with cosine learning rate decay, ensuring stable convergence and effective weight updates. Training is conducted on an NVIDIA RTX A6000 (48GB VRAM), providing the necessary computational power for handling hyperspectral data and deepfake detection tasks efficiently.

\subsection{Summary of HyperFake Pipeline}
HyperFake follows a structured pipeline for deepfake detection. First, the RGB-to-Hyperspectral Reconstruction stage converts standard RGB inputs into 31-channel hyperspectral data using MST++. Next, the Spectral Attention Mechanism selects key spectral bands and refines the extracted features for better deepfake detection. Finally, the Efficient Classification stage employs EfficientNet-B0 to process the spectral features and make accurate real vs. fake predictions.

\section{Training and Evaluation}

\subsection{Datasets}
To assess the performance of HyperFake, we utilize the FaceForensics++ dataset for our preliminary evaluation:

\begin{itemize}
    \item \textbf{FaceForensics++}: This dataset includes 1,000 real videos and their manipulated counterparts, generated using various face manipulation techniques. It offers a comprehensive benchmark for evaluating the performance of deepfake detection algorithms across different manipulation methods. \cite{FaceForensics2019}
\end{itemize}

\begin{figure*}[h]
    \centering
    \includegraphics[width=\textwidth]{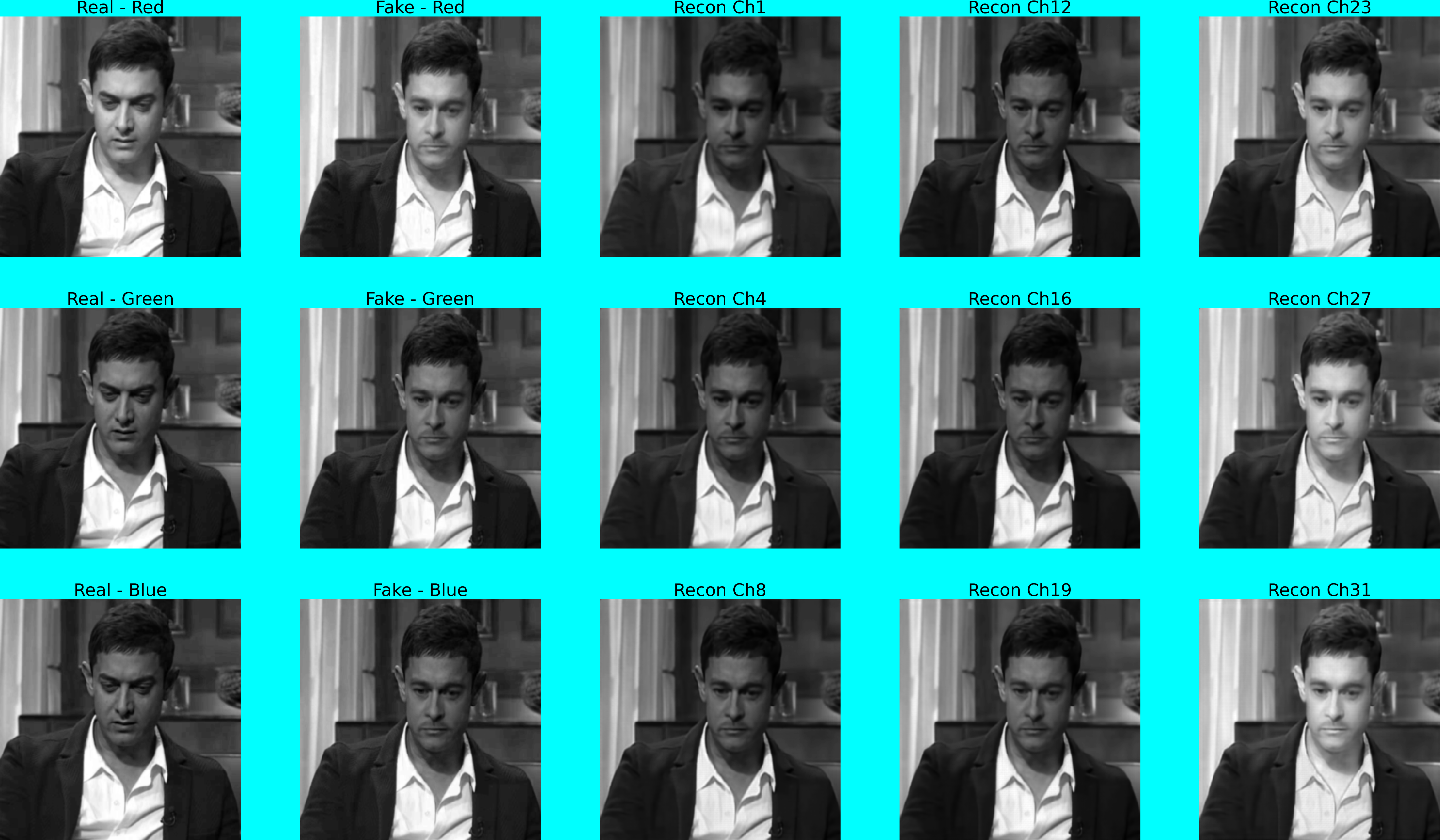}
    \caption{Comparison of real and fake images with hyperspectral reconstructions.The first column shows real images, the second shows fake images, and the last three display selected spectral channels (channels 8, 16, 31) from the reconstructed fakes. These reconstructions highlight subtle differences, helping detect deepfakes.}
    \label{fig:hsi_visualization}
\end{figure*}

\subsection{Evaluation Metrics}
To evaluate the effectiveness of HyperFake, we employ the following metrics:

\begin{itemize}
    \item \textbf{Accuracy}: This metric measures the proportion of correctly classified instances (both real and fake) over the total number of instances. While straightforward, accuracy can be misleading in imbalanced datasets where the number of real and fake videos is not equal.
    
    \item \textbf{Equal Error Rate (EER)}: EER is the point at which the \textbf{False Acceptance Rate (FAR)} equals the \textbf{False Rejection Rate (FRR)}. It provides a single value that represents the trade-off between false positives and false negatives, making it crucial for security-sensitive applications. EER is computed as:

    \begin{equation}
    \text{EER} = \text{FAR}(\tau^*) = \text{FRR}(\tau^*)
    \end{equation}
    
    where \( \tau^* \) is the threshold at which the False Acceptance Rate (FAR) and False Rejection Rate (FRR) are equal. A lower EER indicates a more balanced and accurate model.
    
    \item \textbf{Area Under the Receiver Operating Characteristic Curve (AUC-ROC)}: AUC-ROC measures the model’s ability to distinguish between real and fake samples across all threshold values. It is calculated as:
    
    \begin{equation}
    \text{AUC} = \int_{0}^{1} \text{TPR}(t) \, d\text{FPR}(t)
    \end{equation}
    
    where:
    \begin{itemize}
        \item \( \text{TPR} \) (True Positive Rate) = \( \frac{\text{TP}}{\text{TP} + \text{FN}} \)
        \item \( \text{FPR} \) (False Positive Rate) = \( \frac{\text{FP}}{\text{FP} + \text{TN}} \)
    \end{itemize}
    
    An AUC of 1 indicates perfect classification, while an AUC of 0.5 means the model performs no better than random guessing. This metric is particularly useful for evaluating models trained on imbalanced datasets.  

\end{itemize}

By leveraging these datasets and evaluation metrics, we aim to provide a comprehensive assessment of HyperFake's performance in detecting deepfake content.

\section{Results and Discussion}
We evaluate HyperFake's performance on the FaceForensics++ dataset and compare it with two popular baseline models. The results are shown in Table \ref{tab:results}

\begin{table}[h]
    \centering
    \caption{Model Performance Comparison on FaceForensics++ dataset}
    \label{tab:results}
    \begin{tabular}{lcc}
        \hline
        \textbf{Model} & \textbf{Train Accuracy (\%)} & \textbf{Val Accuracy (\%)} \\
        \hline
        ResNet-50 \cite{resnet} & 72.62 & 63.75 \\
        EfficientNet-B7 \cite{efficientnetb7} & 67.25 & 71.75 \\
        \textbf{HyperFake (Ours)} & \textbf{98.94} & \textbf{92} \\
        \hline
    \end{tabular}
\end{table}

Our experimental results show a clear performance comparison between traditional architectures and our novel HyperFake approach in deepfake detection. Testing on the FaceForensics++ dataset revealed interesting patterns across all three models.

The ResNet-50 architecture showed moderate performance, achieving 72.62\% accuracy during training. However, its validation accuracy dropped to 63.75\%, indicating a significant generalization problem. This gap between training and validation performance suggests that while ResNet-50 can learn patterns from training data, it struggles to apply these learnings to new, unseen examples.

EfficientNet-B7 presented an intriguing case with its unusual performance pattern. It achieved 67.25\% accuracy during training but surprisingly performed better on validation data with 71.75\% accuracy. This inverse relationship between training and validation performance is uncommon and might suggest that the model's regularization techniques are perhaps too aggressive during training.

Our HyperFake model demonstrated remarkable improvements over both baselines. It achieved an impressive 98.94\% accuracy during training while maintaining a strong 92\% accuracy on validation data. This high performance, coupled with only a moderate 6.94\% drop between training and validation, indicates that HyperFake not only learns effectively but also generalizes well to new data. The model's ability to maintain such high accuracy on validation data suggests that its hyperspectral reconstruction approach captures genuinely meaningful features for deepfake detection, rather than just memorizing training examples.

These results validate our hypothesis that incorporating hyperspectral information significantly enhances deepfake detection capabilities compared to traditional RGB-based approaches.

\section{Conclusion}
We introduced HyperFake, a deepfake detection method that uses hyperspectral reconstruction and spectral attention to enhance detection accuracy. By converting standard RGB images into 31-channel hyperspectral data, HyperFake captures subtle spectral inconsistencies that traditional methods miss.

This work presents preliminary results on FaceForensics++. Our approach demonstrates the potential for improving generalization and detecting high-quality deepfakes that evade conventional detectors. However, hyperspectral reconstruction adds computational overhead, making real-time detection challenging.

Future work will focus on four key areas: comprehensive evaluation across DFDC and Celeb-DF datasets, lightweight models for faster inference, extensive cross-dataset evaluation, and architectural optimizations. We plan to explore alternative backbone networks, efficient spectral reconstruction methods, and pruning techniques to reduce model complexity. Additionally, we will investigate multi-modal integration, combining spectral features with audio and temporal information for enhanced detection capabilities.

HyperFake demonstrates the potential of hyperspectral analysis in deepfake detection, offering a new direction for more robust and reliable digital media authentication.

\bibliographystyle{IEEEtran}
\bibliography{references}

@article{fakeout,
  title={FakeOut: Leveraging Out-of-domain Self-supervision for Multi-modal Video Deepfake Detection},
  author={Knafo, Gil and Fried, Ohad},
  journal={arXiv preprint arXiv:2212.00773},
  year={2022},
  url={https://arxiv.org/abs/2212.00773}
}

@inproceedings{blockwise,
  title={Blockwise Spectral Analysis for Deepfake Detection in High-fidelity Videos},
  author={Authors},
  booktitle={Proceedings of the IEEE Conference},
  year={2023},
  url={https://ieeexplore.ieee.org/document/10032370}
}

@article{deepfakeReview,
  title={A Review of Deepfake Techniques: Architecture, Detection, and Datasets},
  author={Authors},
  journal={IEEE Access},
  year={2022},
  url={https://ieeexplore.ieee.org/document/10711187}
}

@inproceedings{FaceForensics2019,
    title={FaceForensics++: Learning to Detect Manipulated Facial Images},
    author={Rössler, Andreas and Cozzolino, Davide and Verdoliva, Luisa and Riess, Christian and Thies, Justus and Nießner, Matthias},
    booktitle={Proceedings of the IEEE/CVF International Conference on Computer Vision},
    pages={1--11},
    year={2019}
}

@inproceedings{MSTPlusPlus2022,
  author    = {Yuanhao Cai and Jing Lin and Zudi Lin and Haoqian Wang and Yulun Zhang and Hanspeter Pfister and Radu Timofte and Luc Van Gool},
  title     = {{MST++}: Multi-stage Spectral-wise Transformer for Efficient Spectral Reconstruction},
  booktitle = {Proceedings of the {NTIRE} Spectral Reconstruction Challenge},
  year      = {2022}
}

@inproceedings{arad1k,
  title={NTIRE 2022 Spectral Recovery Challenge and Data Set},
  author={Arad, Boaz and Timofte, Radu and Yahel, Rony and Morag, Nimrod and Bernat, Amir and Cai, Yuanhao and Lin, Jing and Lin, Zudi and Wang, Haoqian and Zhang, Yulun and Pfister, Hanspeter and Van Gool, Luc and Liu, Shuai and Li, Yongqiang and Feng, Chaoyu and Lei, Lei and Li, Jiaojiao and Du, Songcheng and Wu, Chaoxiong and Leng, Yihong and Song, Rui and Zhang, Mingwei and Song, Chongxing and Zhao, Shuyi and Lang, Zhiqiang and Wei, Wei and Zhang, Lei and Dian, Renwei and Shan, Tianci and Guo, Anjing and Feng, Chengguo and Liu, Jinyang and Agarla, Mirko and Bianco, Simone and Buzzelli, Marco and Celona, Luigi and Schettini, Raimondo and He, Jiang and Xiao, Yi and Xiao, Jiajun and Yuan, Qiangqiang and Li, Jie and Zhang, Liangpei and Kwon, Taesung and Ryu, Dohoon and Bae, Hyokyoung and Yang, Hao-Hsiang and Chang, Hua-En and Huang, Zhi-Kai and Chen, Wei-Ting and Kuo, Sy-Yen and Chen, Junyu and Li, Haiwei and Liu, Song and Sabarinathan, K and Uma, B and Bama, Sathya and Roomi, S Mohamed Mansoor},
  booktitle={Proceedings of the IEEE/CVF Conference on Computer Vision and Pattern Recognition (CVPR) Workshops},
  pages={863--881},
  year={2022}
}

@inproceedings{Coccomini2022,
  title     = {Combining EfficientNet and Vision Transformers for Video Deepfake Detection},
  author    = {Davide Coccomini and Nicola Messina and Claudio Gennaro and Fabrizio Falchi},
  booktitle = {Proceedings of the IEEE/CVF Conference on Computer Vision and Pattern Recognition},
  year      = {2022},
  url       = {https://github.com/davide-coccomini/Combining-EfficientNet-and-Vision-Transformers-for-Video-Deepfake-Detection}
}

@article{Nirkin2020,
  title     = {DeepFake Detection Based on Discrepancies Between Faces and their Context},
  author    = {Yuval Nirkin and Lior Wolf and Yosi Keller and Tal Hassner},
  journal   = {arXiv preprint arXiv:2008.12262},
  year      = {2020},
  url       = {https://arxiv.org/abs/2008.12262}
}

@inproceedings{efficientnet,
  author    = {Mingxing Tan and Quoc V. Le},
  title     = {EfficientNet: Rethinking Model Scaling for Convolutional Neural Networks},
  booktitle = {Proceedings of the 36th International Conference on Machine Learning (ICML)},
  year      = {2019},
  pages     = {6105--6114},
  publisher = {PMLR},
  url       = {https://arxiv.org/abs/1905.11946}
}

@article{laanet,
  title     = {{LAA-Net}: Localized Artifact Attention Network for Quality-Agnostic and Generalizable Deepfake Detection},
  author    = {Dat Nguyen and Nesryne Mejri and Inder Pal Singh and Polina Kuleshova and Marcella Astrid and Anis Kacem and Enjie Ghorbel and Djamila Aouada},
  journal   = {arXiv preprint arXiv:2401.13856},
  year      = {2024},
  url       = {https://arxiv.org/abs/2401.13856}
}

@inproceedings{efficientnetb7,
  title     = {EfficientNet: Rethinking Model Scaling for Convolutional Neural Networks},
  author    = {Mingxing Tan and Quoc V. Le},
  booktitle = {Proceedings of the 36th International Conference on Machine Learning},
  year      = {2019},
  pages     = {6105--6114},
  url       = {https://arxiv.org/abs/1905.11946}
}

@inproceedings{resnet,
  title     = {Deep Residual Learning for Image Recognition},
  author    = {Kaiming He and Xiangyu Zhang and Shaoqing Ren and Jian Sun},
  booktitle = {Proceedings of the IEEE Conference on Computer Vision and Pattern Recognition (CVPR)},
  year      = {2016},
  pages     = {770--778},
  doi       = {10.1109/CVPR.2016.90},
  url       = {https://arxiv.org/abs/1512.03385}
}

\end{document}